\documentclass[graybox]{svmult}

\usepackage{mathptmx}       
\usepackage{helvet}         
\usepackage{courier}        
\usepackage{type1cm}        
\usepackage{makeidx}         
\usepackage{graphicx}        
\usepackage{multicol}        
\usepackage[bottom]{footmisc}

\makeindex             
                       
\usepackage{makecell}
\usepackage{todonotes}
\usepackage{xcolor}
\usepackage{tikz, pgfplots}
\usepackage{graphicx}
\usepackage{booktabs}
\usepackage{amssymb}
\usepackage{amsmath}
\usepackage{multirow}
\usepackage{float}

\usepackage{dashundergaps}

\usepackage{pgf}
\usetikzlibrary{arrows,shapes.misc,chains,scopes}
\usepackage{pgfplotstable}
\pgfplotsset{compat=1.15}

\begin{document}

\title*{Empathetic Dialogue Generation with Pre-trained RoBERTa-GPT2 and External Knowledge}
\titlerunning{Empathetic Dialogue Generation with Pre-trained RoBERTa-GPT2 and EK}

\author{Ye Liu, Wolfgang Maier, Wolfgang Minker \and Stefan Ultes}
\institute{Ye Liu, Wolfgang Maier, Stefan Ultes 
\at Mercedes-Benz AG, Sindelfingen, Germany 
\email{ye.y.liu@daimler.com, wolfgang.mw.maier@daimler.com, stefan.ultes@daimler.com}
\and Wolfgang Minker 
\at Ulm University, Ulm, Germany \email{wolfgang.minker@uni-ulm.de}}
%
%
\maketitle

\abstract*{One challenge for dialogue agents is to recognize feelings of the conversation partner and respond accordingly. In this work, RoBERTa-GPT2 is proposed for empathetic dialogue generation, where the pre-trained auto-encoding RoBERTa is utilised as encoder and the pre-trained auto-regressive GPT-2 as decoder. With the combination of the pre-trained RoBERTa and GPT-2, our model realizes a new state-of-the-art emotion accuracy. To enable the empathetic ability of RoBERTa-GPT2 model, we propose a commonsense knowledge and emotional concepts extractor, in which the commonsensible and emotional concepts of dialogue context are extracted for the GPT-2 decoder. The experiment results demonstrate that the empathetic dialogue generation benefits from both pre-trained encoder-decoder architecture and external knowledge.}

\abstract{One challenge for dialogue agents is to recognize feelings of the conversation partner and respond accordingly. In this work, RoBERTa-GPT2 is proposed for empathetic dialogue generation, where the pre-trained auto-encoding RoBERTa is utilised as encoder and the pre-trained auto-regressive GPT-2 as decoder. With the combination of the pre-trained RoBERTa and GPT-2, our model realizes a new state-of-the-art emotion accuracy. To enable the empathetic ability of RoBERTa-GPT2 model, we propose a commonsense knowledge and emotional concepts extractor, in which the commonsensible and emotional concepts of dialogue context are extracted for the GPT-2 decoder. The experiment results demonstrate that the empathetic dialogue generation benefits from both pre-trained encoder-decoder architecture and external knowledge.}

\section{Introduction}
\label{sec: introduction}

With the development of Spoken Dialogue Systems (SDSs), people are no longer satisfied with the task-oriented interaction, like book a train ticket or make a reservation; but are additionally interested in chit-chat communication. An expected trait of chit-chat agents is to be able to identify the user emotion and express their empathy. For instance, the psychology study in~\cite{zech2005talking} shows that talking about an emotional experience to someone and sharing their emotions contributes to emotional recovery from the event. Hence, exactly identifying the user emotion and appropriately expressing their empathy will be a desired trait for SDSs.


Table \ref{tab: One empathetic dialogue in EmpatheticDialogues dataset.} shows an empathetic dialogue from the EmpatheticDialogues dataset \cite{rashkin2019towards}. A speaker tells a listener the lonely situation that they are facing, and the listener tries to understand the speaker’s feelings and responds accordingly. Even though sharing emotional experiences is a general manifestation for humans, it is a great challenge to train a chit-chat agent capable to understand the user emotion and respond empathetically.

\begin{table}
\caption{One empathetic dialogue in EmpatheticDialogues dataset.}
\label{tab: One empathetic dialogue in EmpatheticDialogues dataset.}

\centering
\begin{tabular}{l c}
\toprule
    \textbf{Emotion} & Lonely  \\
\midrule
\midrule
    \textbf{Situation} & All my friends live in a different country  \\
\midrule
\midrule
    \textbf{Speaker} &  Hi, I feel so lonely sometimes because all my friends live in a different country. \\
    \\
    \textbf{Listener} & \makecell{Oh, I'm sure you are lonely. Maybe you can join some kind of club \\ that lets you meet new friends?} \\
    \\
    \textbf{Speaker} & I was thinking about it! I wanted to join a group for local moms. \\
    \\
    \textbf{Listener} & \makecell{That's a good idea! This way you can also meet friends for yourself, but also maybe \\ meet new friend's  for your  children to hang out with while you do with their moms!}\\
\bottomrule

\end{tabular}
\end{table}

Several works with Transformer-based encoder-decoder architecture \cite{vaswani2017attention} have been presented for empathetic dialogue generation, such as the multi-task learning \cite{rashkin2018know, rashkin2019towards, wei2019emotion} or mixture of experts \cite{lin2019moel}. However, the combination of a pre-trained auto-encoding encoder and a pre-trained auto-regressive decoder have not been explored for empathetic dialogue generation. In this work, the pre-trained Robustly optimized BERT approach (RoBERTa) \cite{DBLP:journals/corr/abs-1907-11692} as encoder and the pre-trained Generative Pre-trained Transformer (GPT-2) \cite{radford2019language} as decoder: RoBERTa-GPT2 encoder-decoder architecture is presented for empathetic dialogue generation. The experiments with EmpatheticDialogues dataset show that the combination of RoBERTa and GPT-2 highly improves the emotion recognition ability and realizes a new state-of-the-art emotion accuracy.

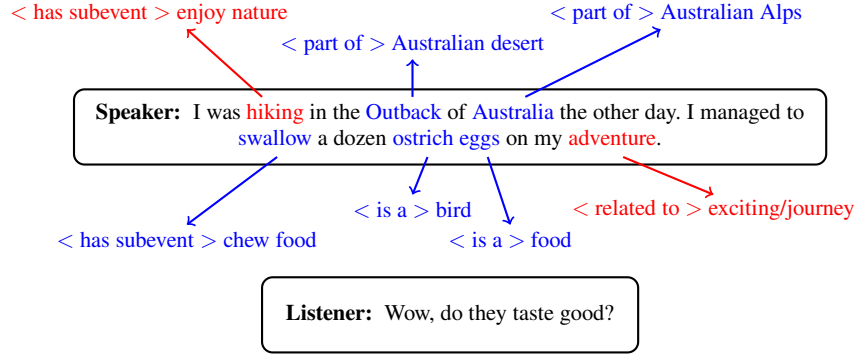
\begin{figure}
\centering
\begin{tikzpicture}[scale=1.0]

\draw[thick, rounded corners]   (-6,-0.5) rectangle (4,0.5);
\node at (-1.0,0)[align=center]  {\textbf{Speaker: } I was \textcolor{red}{hiking} in the \textcolor{blue}{Outback} of \textcolor{blue}{Australia} the other day. I managed to \\ \textcolor{blue}{swallow} a dozen \textcolor{blue}{ostrich} \textcolor{blue}{eggs} on my \textcolor{red}{adventure}.};

\draw[thick, rounded corners]   (-3.5,-3.0) rectangle (1.5,-2.0);
\node at (-1.0,-2.5)[align=center]  {\textbf{Listener: } Wow, do they taste good?};

\node at (-5,1.5) {\textcolor{red}{$<$ has subevent $>$ enjoy nature}};
\node at (-1.5,1.1) {\textcolor{blue}{$<$ part of $>$ Australian desert}};
\node at (2.0,1.5) {\textcolor{blue}{$<$ part of $>$ Australian Alps}};

\draw[->,thick,red] (-3.5,0.4) -- (-4.5,1.3);
\draw[->,thick,blue] (-1.5,0.4) -- (-1.5,0.9);
\draw[->,thick,blue] (0.0,0.4) -- (1.8,1.3);

\node at (-4.5,-1.5) {\textcolor{blue}{$<$ has subevent $>$ chew food}};
\node at (-1.5,-1.1) {\textcolor{blue}{$<$ is a $>$ bird}};
\node at (-0.2,-1.5) {\textcolor{blue}{$<$ is a $>$ food}};
\node at (2.5,-1.1) {\textcolor{red}{$<$ related to $>$ exciting/journey}};

\draw[->,thick,blue] (-3.3,-0.4) -- (-4.5,-1.3);
\draw[->,thick,blue] (-1.3,-0.4) -- (-1.5,-0.9);
\draw[->,thick,blue] (-0.5,-0.4) -- (-0.2,-1.3);
\draw[->,thick,red] (1.3,-0.4) -- (2.5,-0.9);

\end{tikzpicture}
\caption{\label{fig: an example of EmpatheticDialogues dataset with underlying commonsense knowledge and emotional concepts.} An example of EmpatheticDialogues dataset with underlying commonsense knowledge (blue part) and emotional concepts (red part). (The special token in $<$ $>$ represents the relation in commonsense knowledge: ConcepetNet \cite{speer2017conceptnet}.)}
\end{figure}

In addition to the advanced neural network architecture, some external knowledge also contributes to the empathetic dialogue generation. Humans generally understand the world and express implicit emotions based on their experience and knowledge. Also, \cite{young2018augmenting} demonstrates that commonsense knowledge is fundamental for chit-chat agents to understand conversations and generate appropriate responses. As shown in Fig. \ref{fig: an example of EmpatheticDialogues dataset with underlying commonsense knowledge and emotional concepts.}, the underlying commonsensible and emotional concepts of the speaker utterance can help the listener to better understand what the speaker is talking about. Hence, we propose an Commonsense Knowledge and Emotional Concepts Extractor (CKECE) for GPT-2 decoder in our work, to enable the commonsense and empathetic response generation. In the CKECE, we firstly utilize KeyBERT \cite{grootendorst2020keybert} to extract the keywords from the dialogue context; then elicit the commonsensible and emotional concepts of the keywords based on commonsense knowledge: ConceptNet \cite{speer2017conceptnet} and emotion lexicon: NRC\_VAD \cite{mohammad2018obtaining}; finally the extracted concepts are fed into GPT-2 decoder in a more plain text format to guide the empathetic generation.

\section{Related Work}
\label{sec: realted work}
Open-domain and chit-chat conversational models have been widely studied \cite{serban2016generative, wolf2019transfertransfo}. With the rise of public accessible datasets \cite{hsu2018emotionlines, li2017dailydialog, rashkin2019towards} and data-driven learning approaches \cite{sutskever2014sequence, vaswani2017attention}, several works have attempted to make chit-chat dialogue more engaging. Some aim to improve the personalization of responses by conditioning the generation on a persona profile \cite{li2016persona}. Then the PersonaChat dataset \cite{zhang2018personalizing} was particularly introduced and the competition in ConvAI $2$ challenge \cite{dinan2019second} demonstrated that the produced responses include more consistent personas by adding persona information into the model. However, the personalized dialogue models often can not take the feelings of their conversation partners into consideration. Besides the chit-chat research on displaying the consistent personality, some works focus on emotional and empathetic dialogue generation. The existing emotional dialogue models \cite{colombo2019affect, li2018syntactically, shen2020cdl, zhou2018emotional, zhou2018mojitalk} generally generate the response depending on a predefined emotion, however, the empathetic dialogue models are capable of perceiving the emotion of the speaker and express their empathy without extra step to determine which emotion type to respond explicitly \cite{skowron2013affect}. Hence, the empathetic dialogue model is more in line with the real world scenarios \cite{li2020empathetic}, because the listener is capable to infer the emotion of the speaker in human-human communication.


In recent years, several works have been presented for empathetic dialogue generation. \cite{rashkin2019towards} created a benchmark and dataset towards empathetic open-domain dialogue. \cite{lin2019moel} softly combined the possible emotional responses from several separate experts to generate the final empathetic response. \cite{li2020empdg} proposed an multi-resolution interactive empathetic dialogue model to evoke more emotional perception in dialogue generation. \cite{li2020empathetic} proposed a multi-type knowledge aware empathetic dialogue generation framework to enhance the empathy of generations. The above-mentioned approaches are all trained from scratch. \cite{naous2021empathetic} proposed BERT2BERT for Arabic empathetic response generation, while the encoder and decoder are both warm started using pre-trained auto-encoding AraBERT \cite{antoun2020arabert} parameters. \cite{zandie2020emptransfo} introduced EmpTransfo and \cite{lin2020caire} presented CAiRE, both are empathetic aware model adapted from GPT \cite{radford2018improving}. With the release of encoder-decoder model in Huggingface\footnote{https://huggingface.co/transformers/model\_doc/encoderdecoder.html}, where any pre-trained auto-encoding model as the encoder and any pre-trained auto-regressive model as the decoder can be initialized as a sequence-to-sequence model, we are more interested in the performance of pre-trained auto-encoding encoder and auto-regressive decoder architecture for empathetic dialogue generation. Furthermore, \cite{rothe2020leveraging} performed an extensive study on leveraging variable pre-trained models for sequence generation tasks and demonstrated that combining RoBERTa \cite{DBLP:journals/corr/abs-1907-11692} and GPT-2 \cite{radford2019language} achieves strong results. Hence, RoBERTa-GPT2 is proposed in this work for empathetic dialogue generation. 


In addition, the corpora with emotion labelling play a significant role in empathetic dialogue generation. There are several interesting resources. \cite{li2017dailydialog} developed the DailyDialog dataset, with manually emotion labelling to each utterance. \cite{hsu2018emotionlines} collected the EmotionLines dataset from TV shows and human-to-human chats, where each utterance is further annotated with one of seven emotion-categorical labels. However, only $5\%$ of the utterances in DailyDialog and $16.68\%$ in EmotionLines have varied emotional labels and others are either ``none'' or ``happy'' labels. Hence, they are not suitable for empathetic dialogue generation because of the extremely unbalanced data distribution. \cite{rashkin2019towards} released an empathetic dialogue dataset: EmpatheticDialogues, which focuses explicitly on conversations about emotionally grounded personal situations and considers a richer, evenly distributed set of emotions. In our work, we conduct the experiment of empathetic dialogue generation with EmpatheticDialogues dataset.

\section{The Proposed Method}
\label{sec: the proposed method}
In this work, we present the RoBERTa-GPT2 encoder-decoder architecture for empathetic dialogue generation, where the pre-trained auto-encoding RoBERTa \cite{DBLP:journals/corr/abs-1907-11692} as encoder and pre-trained auto-regressive GPT-2 \cite{radford2019language} as decoder. In addition, a Commonsense Knowledge and Emotional Concepts Extractor (CKECE), which is used to extract the relevant concepts from dialogue history, is proposed to enable the commonsensible and empathetic ability of GPT-2 decoder. In this section, the CKECE will be firstly introduced and then the RoBERTa-GPT2 architecture with extracted concepts will be shown.

\subsection{Commonsense Knowledge and Emotional Concepts Extractor: \textbf{CKECE}}
\label{subsec: emotional concept filter}
For the CKECE, two knowledge sources: the commonsense knowledge ConceptNet \cite{speer2017conceptnet} and the emotional lexicon NRC\_VAD \cite{mohammad2018obtaining}, and one keyword extraction tool, KeyBERT \cite{grootendorst2020keybert}, are used. We firstly utilize the KeyBERT to extract the keywords of the dialogue context, and then filter out the most relevant commonsense knowledge and emotional concepts of the keywords with the confidence score of ConceptNet and emotional intensity of NRC\_VAD.

\subsubsection{The CKECE components}
\label{subsubsec: the preliminaries of CKECE}
The three resources used in CKECE are introduced in the following:

\textbf{KeyBERT}\footnote{https://github.com/MaartenGr/KeyBERT} is a minimal and easy-to-use keyword extraction technique that leverages BERT embeddings and cosine similarity to find the keywords and keyphrases in a document that are the most similar to the document itself.

\textbf{ConceptNet}\footnote{https://conceptnet.io/} is a large-scale and multilingual commonsense knowledge graph that describes general human knowledge in natural language. It comprises $5.9$M assertions, $3.1$M concepts and $38$ relations. The nodes in ConceptNet are concepts and the edges are relations. Each $(\emph{concept1, relation, concept2})$ triplet is an assertion. Each assertion is associated with a confidence score. The assertion confidence score are usually in the $[1, 10]$ interval. For example, $(\emph{loneliness, CausesDesire, socialize})$ with confidence score of $3.464$.

\textbf{NRC\_VAD}\footnote{https://saifmohammad.com/WebPages/nrc-vad.html} is a lexicon that includes a list of more than $20$k English words and their Valence, Arousal, and Dominance (VAD) scores. For a given word and a dimension, the scores range from $0$ (lowest) to $1$ (highest). The interpretations of NRC\_VAD dimensions  are presented in Table \ref{tab: Interpretations of NRC VAD dimensions.}. Such as, the VAD score vector $[V_{a}, A_{r}, D_{o}]$ of word ``happiness'' is [0.960, 0.732, 0.850].

\begin{table}
\caption{Interpretations of NRC\_VAD dimensions.}
\label{tab: Interpretations of NRC VAD dimensions.}

\centering
\setlength{\tabcolsep}{6pt}
\begin{tabular}{ccc}
\toprule
    Dimensions & Values & Interpretations \\
\midrule
    Valence ($V_{a}$) & [0, 1] & Negative-Positive \\
    Arousal ($A_{r}$) & [0, 1] & Calm-Excited \\
    Dominance ($D_{o}$) & [0, 1] & Weak-Powerful \\
\bottomrule

\end{tabular}
\end{table}

\subsubsection{CKECE}
\label{subsubsec: CKECE}
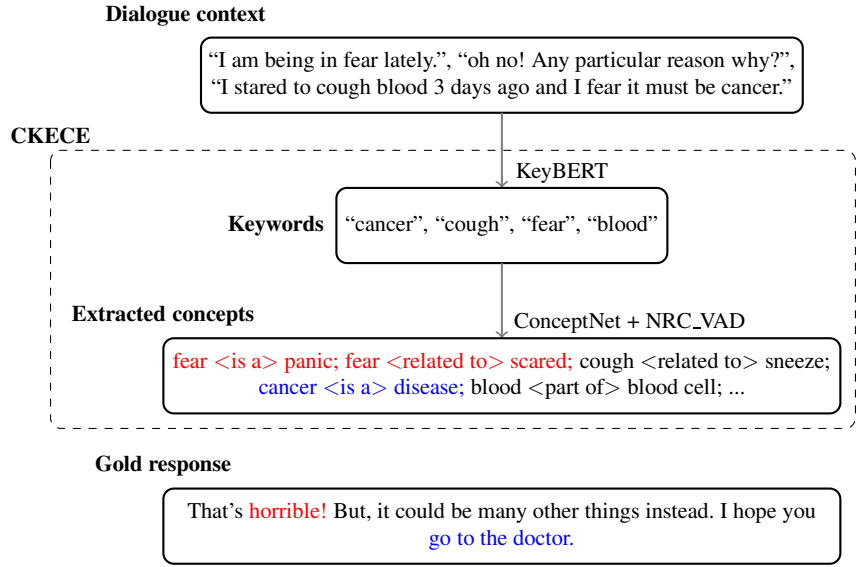
\begin{figure}
\centering
\begin{tikzpicture}[scale=1]

\node at (-5.2,0.8) {\textbf{Dialogue context}};
\draw[thick, rounded corners]   (-5,-0.5) rectangle (3,0.5);
\node at (-1.0,0)[align=center]  { ``I am being in fear lately.'', ``oh no! Any particular reason why?'', \\ ``I stared to cough blood 3 days ago and I fear it must be cancer.''};

\draw[->,thick,gray] (-1.0,-0.5) -- (-1.0,-1.5);
\node at (-0.2,-1.3) {KeyBERT};

\draw[dashed, rounded corners] (-7.0,-1.0) rectangle (3.7,-4.7);
\node at (-7.0,-0.8) {\textbf{CKECE}};

\node at (-4,-2) {\textbf{Keywords}};
\draw[thick, rounded corners]   (-3.2,-2.5) rectangle (1.2,-1.5);
\node at (-1.0,-2)[align=center] {``cancer'', ``cough'', ``fear'', ``blood''};

\draw[->,thick,gray] (-1.0,-2.5) -- (-1.0,-3.5);
\node at (0.7,-3.3) {ConceptNet + NRC\_VAD};

\node at (-5.5,-3.2) {\textbf{Extracted concepts}};
\draw[thick, rounded corners]   (-5.5,-3.5) rectangle (3.5,-4.5);
\node at (-1.0,-4)[align=center] {\textcolor{red}{fear $<$is a$>$ panic;  fear $<$related to$>$ scared;} cough $<$related to$>$ sneeze; \\ \textcolor{blue}{cancer $<$is a$>$ disease;} blood $<$part of$>$ blood cell; ...};

\node at (-5.5,-5.2) {\textbf{Gold response}};
\draw[thick, rounded corners]   (-5.5,-5.5) rectangle (3.5,-6.5);
\node at (-1.0,-6)[align=center] {That's \textcolor{red}{horrible!} But, it could be many other things instead. I hope you \\ \textcolor{blue}{go to the doctor.}};

\end{tikzpicture}
\caption{\label{fig: an example for the process of commonsense knowledge and emotional concepts extraction.} An example for the process of CKECE for the dialogue context. The extracted emotional concepts and emotional word in gold response are marked in red. The blue part in extracted concepts and gold response share same commonsense knowledge.}
\end{figure}

To extract more relevant concepts, we firstly utilize the KeyBERT to extract the keywords from the dialogue context. In this step, the recommended KeyBERT model ``distilbert-base-nli-mean-tokens'' is used and only maximal top $10$ keywords with score larger than $0$ are retained. An example of extracted keywords is shown in Fig. \ref{fig: an example for the process of commonsense knowledge and emotional concepts extraction.}.

Then, we pick out the commonsense concepts from ConceptNet based on the keywords and denote them in a tuple (keyword, relation, concept, scaled confidence score) as $\{\tau_{k}^{i} = (k_{i}, r_{k}^{i}, c_{k}^{i}, s_{k}^{i})\}_{k=1,2,\dots,K}$ where the confidence score $s$ is scaled by the following Eq. \ref{equ: min-max of conceptnet} $\emph{min-max}$ normalization.

\begin{equation}
\label{equ: min-max of conceptnet}
    \emph{min-max}(s) = \frac{s-min_{s}}{max_{s}-min_{s}} \; ,
\end{equation}
where $min_{s}$ is $1$ and $max_{s}$ is $10$. The processed $s \in [0, 1]$ and the $\emph{min-max}$ normalization is also used in \cite{li2020empathetic, zhong2019knowledge}. With $\emph{min-max}$ normalization, the example $(\emph{loneliness, CausesDesire, socialize})$ with confidence score $3.464$ in Section \ref{subsubsec: the preliminaries of CKECE} is transformed into (loneliness, CausesDesire, socialize, $0.274$) tuple with scaled confidence score $0.274$. In order to pick out the most relevant concepts, the following tuples will be removed in this step:
\begin{itemize}
\item{The keywords or concepts are stop words. (The union of stop words in NLTK \cite{loper2002nltk} and SpaCy\footnote{https://github.com/explosion/spaCy} are used.)}
\item{The scaled confidence score is less than a pre-defined threshold $\alpha$. We set $\alpha$ is $0.1$ in this work, i.e. $s < 0.1$.}
\item{The keywords and concepts are same or have the same stem. Like: (addition, Synonym, addition, $0.11$); (actual, DerivedFrom, actually, $0.11$).}
\item{The relation is in an excluded relation list. i.e. $r \in [Antonym, ExternalURL, \\ NotDesires, NotHasProperty, NotCapableOf, dbpedia, DistinctFrom, \\ EtymologicallyDerivedFrom, EtymologicallyRelatedTo, SymbolOf, FormOf, \\ AtLocation, DerivedFrom, SymbolOf]$}
\end{itemize}

Furthermore, to enable the emotional concepts, we adopt NRC\_VAD to compute emotion intensity values for the concepts $c$ as the Eq. \ref{equ: emotion intensity}.

\begin{equation}
\label{equ: emotion intensity}
\eta({c}) = \emph{min-max}(	\| V_{a}(c)-\frac{1}{2}, \frac{A_{r}(c)}{2}	\|_{2}) \; ,
\end{equation}
where $\| \cdot \|_{k}$ denotes $l_{k}$ norm. $V_{a}(c)$ and $A_{r}(c)$ represent valence and arousal score of concept $c_{i}$, respectively. When $c$ not in NRC\_VAD, we set $\eta({c})$ to the mid value of $0.5$. 


Lastly, the final score $f$ in Eq. \ref{equ: final score} is derived from three aspects: emotion intensity, semantic similarity and scaled confidence score. The semantic similarity $cos(k_{i}, c_{k}^{i})$ is the cosine similarity between keyword and concept both embedded by the GloVe \cite{pennington2014glove}, which stands for global vectors for word representation and is an unsupervised learning algorithm for obtaining vector representations for words.

\begin{equation}
\label{equ: final score}
f(\tau_{k}^{i}) = \eta({c_{k}^{i}}) + cos(k_{i}, c_{k}^{i}) + s_{k}^{i} \; ,
\end{equation}
We sort the candidate tuples in descending order of the final scores and select top $3$ tuples for each keyword. Maximal $10$ tuples are chosen for every dialogue context. Then the extracted concepts are arranged in a more plain textual form: ``keyword $<$relation$>$ concept'', which is shown in Fig. \ref{fig: an example for the process of commonsense knowledge and emotional concepts extraction.}, for GPT-2 decoder.

\subsection{Pre-trained RoBERTa-GPT2 encoder-decoder}
\label{subsec: RoBERTa-GPT2}
The RoBERTa \cite{DBLP:journals/corr/abs-1907-11692} and GPT-2 \cite{radford2019language} are both large architectures pre-trained on
large collections of texts. Then the pre-trained models are widely fine-tuned in downstream tasks. In this work, we explore the pre-trained RoBERTa-GPT2 as encoder-decoder architecture for empathetic dialogue generation.

\subsubsection{The preliminaries of RoBERTa-GPT2}
\label{subsubsec: the preliminaries of RoBERTa-GPT2}
The pre-trained auto-encoding RoBERTa and pre-trained auto-regressive GPT-2 are introduced in the following:

\textbf{RoBERTa}\footnote{https://github.com/pytorch/fairseq/tree/master/examples/roberta} has the same architecture as BERT \cite{devlin2019bert}, but uses a byte-level Byte-Pair Encoding (BPE) \cite{sennrich2016neural} as a tokenizer (same as GPT-2) and improved the training procedure of BERT \cite{devlin2019bert}. 


\textbf{GPT-2}\footnote{https://github.com/openai/gpt-2} is a pre-trained large-scale unsupervised language model which generates coherent paragraphs of text. GPT-2 is also widely used in task-oriented dialogue generation \cite{ budzianowski2019hello, peng2020few} and chit-chat dialogue generation \cite{lin2020caire, zhang2020dialogpt}. 


\subsubsection{RoBERTa-GPT2}
\label{subsubsec: roberta-gpt2}
\begin{figure}
\centering
\begin{tikzpicture}[scale=1]

\draw[thick,rounded corners]   (-5.5,-0.5) rectangle (-0.5,0.5);
\node at (-3.0,0)[align=center]  {RoBERTa};

\draw[->,thick] (-3.0,-1) -- (-3.0,-0.5);
\node at (-1.5,-1.2) {$<$CLS$>$ i am ... $<$SEP$>$ oh no! ... $<$SEP$>$ i started ... $<$SEP$>$};

\draw[dashed, ->,thick] (1.0,-1) -- (1.0,-0.5);
\draw[dashed, thick,rounded corners]  (0.3,-0.4) rectangle (1.5,0.2);
\node at (0.9,-0.1)[align=center]  {CKECE};
\draw[dashed, ->,thick] (1.0,0.2) -- (1.0,0.7);
\draw [thick,dash dot] (-0.5, 0.8) -- (2.0, 0.8);

\draw[thick,rounded corners]   (-5.5,1.5) rectangle (-4.3,2.5);
\node at (-4.9,2)[align=center]  {emotion \\ classifier};

\draw[->,thick] (-5.0,2.5) -- (-5.0,3);
\node at (-4.6,3.1) {emotion logits};

\draw[->,thick] (-5.0,0.5) -- (-5.0,1.5);
\node at (-4.5,0.8)[align=center] {pooled \\ output};

\draw[->,thick,rounded corners] (-1.0,0.5) -- (-1.0,2) -- (1,2);
\node at (-1.5,0.8)[align=center] {encoder \\ outputs};

\draw[thick,rounded corners]   (1,1.5) rectangle (6,2.5);
\node at (3.5,2)[align=center]  {GPT-2};

\draw[->,thick] (4.0,1.0) -- (4.0,1.5);
\node at (2.8, 0.9)[align=center]  {fear $<$is a$>$ panic, ...$<$SEP$>$ that's horrible, ... $<$END$>$};

\draw[->,thick] (4.0,2.5) -- (4.0,3);
\node at (4.5,3.1)[align=center]  {that's horrible, ... $<$END$>$};

\end{tikzpicture}
\caption{\label{fig: the proposed RoBERTa-GPT2 encoder-dcoder architecture with CKECE guidance for empathetic dialogue generation} Our proposed RoBERTa-GPT2 encoder-decoder architecture with CKECE guidance for empathetic dialogue generation.}
\end{figure}
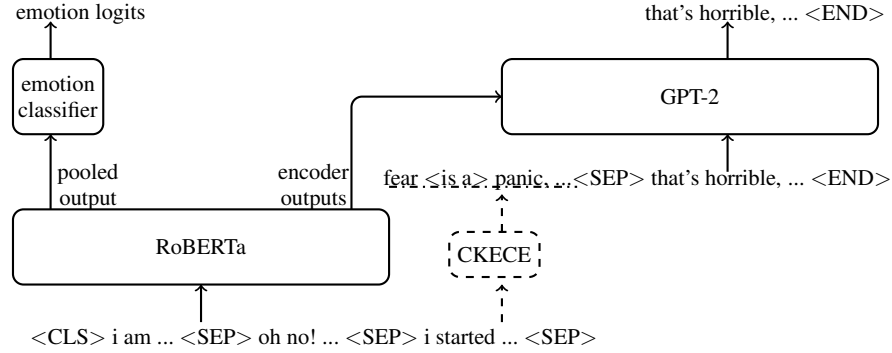

Fig. \ref{fig: the proposed RoBERTa-GPT2 encoder-dcoder architecture with CKECE guidance for empathetic dialogue generation} shows our proposed RoBERTa-GPT2 encoder-decoder architecture for empathetic dialogue generation. The simplified input for RoBERTa encoder and GPT-2 decoder in Fig. \ref{fig: the proposed RoBERTa-GPT2 encoder-dcoder architecture with CKECE guidance for empathetic dialogue generation} only shows the initial part of the sentences. And Fig. \ref{fig: an example for the process of commonsense knowledge and emotional concepts extraction.} and Fig. \ref{fig: the proposed RoBERTa-GPT2 encoder-dcoder architecture with CKECE guidance for empathetic dialogue generation} share the same dialogue example.

The pre-trained RoBERTa as encoder process the dialogue context, where the $<\mathrm{CLS}>$ token is appended at the first place and $<\mathrm{SEP}>$ is for separating speaker utterance and listener utterance. The output of $<\mathrm{CLS}>$ token, pooled output, represents the entire meaning of the input. A linear layer with softmax activation is added on the top of pooled output for emotion classification. The encoder outputs will be fed to the GPT-2 decoder for cross-attention mechanism. As shown in Fig. \ref{fig: the proposed RoBERTa-GPT2 encoder-dcoder architecture with CKECE guidance for empathetic dialogue generation}, the input for GPT-2 decoder starts with extracted concepts. During the training, the gold response is also attached after concepts for faster convergence and separated by $<$SEP$>$ token. It is noteworthy that only the response part without extracted concepts is the output of GPT-2 decoder for computing the generation loss during the training. That means, the response is generated conditioned on the contextual information of encoder outputs with cross-attention mechanism and emotional concepts of decoder inputs with self-attention mechanism by combining pre-trained RoBERTa and GPT-2. Lastly, all the parameters of RoBERTa-GPT2 are jointly trained end-to-end to optimize the emotion classification and response generation by minimising emotion cross entropy loss and maximum likelihood estimator (MLE) generation loss.


\section{Experimental Settings and Results Analysis}
\label{sec: experimental settings and evaluation}

\subsection{Dataset}
\label{subsec: dataset}
We conduct our experiment on the large-scale multi-turn EmpatheticDialogues \cite{rashkin2019towards}, which consists of $25$k one-to-one open-domain conversation grounded in emotional situations. And the EmpatheticDialogues dataset provides $32$ evenly distributed emotion labels.

\subsection{Baselines}
\label{subsec: baselines}
We compare our models with the following four baselines.
\begin{itemize}
    \item [1)]
    {\textbf{Transformer} \cite{vaswani2017attention}: a Transformer-based encoder-decoder model trained with MLE generation loss.}
    
    \item [2)]
    {\textbf{EmoPrepend-1} \cite{rashkin2019towards}: an extension of Transformer model with an additional supervised emotion classifier. The whole model is jointly trained by optimizing both the classification and generation loss.}
    
    \item [3)]
    {\textbf{MoEL} \cite{lin2019moel}: another extension of Transformer model, which softly combines the outputs of the multiple listeners. Each listener is optimized to react to a certain emotion and generate an empathetic response.}
    
    \item [4)]
    {\textbf{MK-EDG} \cite{li2020empathetic}: a multi-type knowledge aware empathetic dialogue generation framework. Commonsense knowledge and emotional lexicon are used to enrich the dialogue utterance.}
\end{itemize}
Additionally, to better analyse our proposed RoBERTa-GPT architecture for empathetic dialogue model, we also conducted \textbf{RoBERTa w/o GPT-2}: only RoBERTa encoder with emotion classifier trained with emotion loss; and \textbf{RoBERTa-GPT2 w/o CKECE}: RoBERTa-GPT2 without the guidance of external knowledge.

\subsection{Training details}
\label{subsec: training details}
The RoBERTa-GPT2 is trained with batch size $16$ and learning rate $1\mathrm{e}{-5}$. Early stopping is applied during the training for saving the best model. During decoding, we use the top-k~\cite{fan2018hierarchical} and nucleus sampling (top-p)~\cite{holtzman2019curious} decoding algorithms with top-k equal to $5$ and top-p equal to $0.9$.

\subsection{Automatic Evaluation Results}
\label{subsec: automatic evaluation results}
To evaluate the performance of RoBERTa-GPT2 model, we firstly adopt the Emotion Accuracy as the agreement between the ground truth emotion labels and the predicted emotion labels by the emotion classifier. In addition, Perplexity \cite{serban2015hierarchical} values are utilized to measure the high-level general quality of the generation model. Furthermore, Distinct-1 and Distinct-2 \cite{li2016diversity} are used to measure the proportion of the distinct unigrams and bigrams in all the generated results to indicate diversity. Table \ref{tab: Evaluation results between RoBERTa-GPT and baselines} shows the evaluation results between our proposed methods and baselines. The results of MK-EDG in Table \ref{tab: Evaluation results between RoBERTa-GPT and baselines} are directly copied from \cite{li2020empathetic}, hence MK-EDG is absent from use cases in Table \ref{tab: generated responses from Transformer, EmoPrepend-1, MoEL and RoBERTa-GPT2}.

\begin{table}
\caption{Evaluation results between RoBERTa-GPT2 and baselines}
\label{tab: Evaluation results between RoBERTa-GPT and baselines}
\centering
\setlength{\tabcolsep}{5pt}
\begin{tabular}{lcccc}
\toprule
    Models & Emotion Accuracy & Perplexity & Distinct-1 & Distinct-2 \\
\midrule
    Transformer & - & 35.56 & 0.41 & 1.49 \\
    EmoPrepend-1 & 0.3359  & 35.66 & 0.42 & 1.62 \\
    MoEL & 0.3425 & 37.69 & 0.43 & 1.72 \\
    MK-EDG  & 0.3931 & 34.85 & 1.48 & 4.90 \\
\midrule
    RoBERTa w/o GPT-2 & 0.3439 & - & - & - \\
    RoBERTa-GPT2 w/o CKECE & \textbf{0.5262} & 14.97 & 1.62 & 10.47 \\
    RoBERTa-GPT2 & 0.5151 & \textbf{13.57} & \textbf{2.04} & \textbf{11.68} \\
\bottomrule
\end{tabular}
\end{table}

In Table \ref{tab: Evaluation results between RoBERTa-GPT and baselines}, we observe our proposed RoBERTa-GPT2 outperforms the baselines by a large margin in terms of all automatic metrics and realizes a new state-of-the-art result. The emotion accuracy comparison between only RoBERTa encoder w/o GPT-2 and RoBERTa-GPT2 demonstrates that emotion classification benefits from the combination of RoBERTa encoder and GPT-2 decoder architecture. The RoBERTa-GPT2 has better results on Perplexity, Distinct-1 and Distinct-2 than RoBERTa-GPT2 w/o CKECE, indicating that injecting external knowledge contributes to the generation quality, though a little loss on emotion accuracy.

\subsection{Use cases study}
\label{subsec: Use cases study}

\begin{table}[h!]
\caption{Generated responses from Transformer, EmoPrepend-1, MoEL and RoBERTa-GPT2. The \underline{underline} part shows the empathetic and contextual of RoBERTa-GPT2 response.}
\label{tab: generated responses from Transformer, EmoPrepend-1, MoEL and RoBERTa-GPT2}

\centering
\begin{tabular}{l c}
\toprule
    Emotion & excited  \\
    Dialogue context & \makecell[c]{\textbf{Speaker:} I'm taking a vacation and going on a cruise \\ to Alaska next week!} \\
\midrule
    Transformer & Where are you going? \\
    EmoPrepend-1 &  That is awesome! Where are you going?\\
    MoEL  & I hope you have a great time!\\
    RoBERTa-GPT2 & I am sure you will \underline{enjoy the cruise!}\\
\midrule
\midrule
    Emotion & terrified  \\
    Dialogue context & \makecell[c]{\textbf{Speaker:} Do you even know how crazy it is to skydive? I must have been \\ absolutely nuts when I did it. \\
    \textbf{Listener:} Oh I don't think I could do that. I have a fear of falling \\ from high places. \\
    \textbf{Speaker:} It gave me the biggest rush that's for sure. But on the  way \\ down I was saying my prayers.} \\
\midrule
    Transformer & That is a great idea. I have been there. \\
    EmoPrepend-1 &  I would be too. I do not know what you mean. \\
    MoEL  & I am sorry to hear that. I hope everything works out for you.\\
    RoBERTa-GPT2 & That is a lot of \underline{stress!} \underline{I am glad you did not get injured.} \\
\midrule
\midrule
    Emotion & excited  \\
    Dialogue context & \makecell[c]{\textbf{Speaker:} I couldn't wait to go to the concert.} \\
\midrule
    Transformer & That sounds like a lot of fun! \\
    EmoPrepend-1 &  That sounds like a lot of fun!\\
    MoEL  & That sounds like a lot of fun!\\
    RoBERTa-GPT2 & That is \underline{awesome}! \underline{What kind of concert will it be?}\\
\midrule
\midrule
    Emotion & trusting  \\
    Dialogue context & \makecell[c]{\textbf{Speaker:} I let my daughter ride in the road by herself on her \\ skateboard the other day.} \\
\midrule
    Transformer & That is awesome! What was it? \\
    EmoPrepend-1 &  That must have been a lot of fun! \\
    MoEL  & oh no! What happened?\\
    RoBERTa-GPT2 & \underline{Wow! I would be a little nervous if she did not skate at all.} \\
\bottomrule
\end{tabular}
\end{table}

Cases of generated responses from baselines Transformer, EmoPrepend-1, MoEL and our proposed method, RoBERTa-GPT2, are listed in Table \ref{tab: generated responses from Transformer, EmoPrepend-1, MoEL and RoBERTa-GPT2}. In the first case, Transformer and EmoPrepend-1 do not exactly understand what the speaker is saying. MoEL identifies the user emotion but the response is kind of generic. Besides the correct understanding user emotion, RoBERTa-GPT2 also knows that the speaker is talking about ``cruise''. The baselines in the second case do not correctly recognise the user emotion. Compared with generic response of the baselines in the third case, RoBERTa-GPT2 generates contextual response with a proper positive emotion by replying with ``awesome''. In the fourth case, the response of EmoPrepend-1 is generic and other two baselines do not understand the speaker, while RoBERTa-GPT2 generates coherent and informative response by showing the concern. All the cases in Table \ref{tab: generated responses from Transformer, EmoPrepend-1, MoEL and RoBERTa-GPT2} show that our proposed RoBERTa-GPT2 can both handle with user emotion and dialogue content.

\section{Conclusion and Outlook}
\label{sec: conclusion and outlook}

In this work, we leverage pre-trained auto-encoding RoBERTa as encoder and pre-trained auto-regressive GPT-2 as decoder for empathetic dialogue generation. Meanwhile, the external knowledge: commonsense knowledge and emotional lexicon; are utilized to extract emotional and commonsensible concepts from  dialogue context for GPT-2 decoder to enable the empathetic and contextual responses. Both automatic metrics and cases study show that our proposed RoBERTa-GPT2 outperforms the baselines and demonstrate that the empathetic dialogue generation benefits from pre-trained modelling and external knowledge.

In the future work, we will continually evaluate our proposed method for empathetic dialogue generation from human perspective. Meanwhile, we are also interested in other flexible methods for injecting external knowledge to empathetic dialogue system.

\bibliographystyle{spmpsci}
\bibliography{iwsds}

\begin{thebibliography}{10}
\providecommand{\url}[1]{{#1}}
\providecommand{\urlprefix}{URL }
\expandafter\ifx\csname urlstyle\endcsname\relax
  \providecommand{\doi}[1]{DOI~\discretionary{}{}{}#1}\else
  \providecommand{\doi}{DOI~\discretionary{}{}{}\begingroup
  \urlstyle{rm}\Url}\fi

\bibitem{antoun2020arabert}
Antoun, W., Baly, F., Hajj, H.: Arabert: Transformer-based model for arabic
  language understanding.
\newblock In: LREC 2020 Workshop Language Resources and Evaluation Conference
  11--16 May 2020, p.~9

\bibitem{budzianowski2019hello}
Budzianowski, P., Vuli{\'c}, I.: Hello, it’s gpt-2-how can i help you?
  towards the use of pretrained language models for task-oriented dialogue
  systems.
\newblock In: Proceedings of the 3rd Workshop on Neural Generation and
  Translation, pp. 15--22 (2019)

\bibitem{colombo2019affect}
Colombo, P., Witon, W., Modi, A., Kennedy, J., Kapadia, M.: Affect-driven
  dialog generation.
\newblock In: Proceedings of the 2019 Conference of the North American Chapter
  of the Association for Computational Linguistics: Human Language
  Technologies, Volume 1 (Long and Short Papers), pp. 3734--3743 (2019)

\bibitem{devlin2019bert}
Devlin, J., Chang, M.W., Lee, K., Toutanova, K.: Bert: Pre-training of deep
  bidirectional transformers for language understanding.
\newblock In: Proceedings of the 2019 Conference of the North American Chapter
  of the Association for Computational Linguistics: Human Language
  Technologies, Volume 1 (Long and Short Papers), pp. 4171--4186 (2019)

\bibitem{dinan2019second}
Dinan, E., Logacheva, V., Malykh, V., Miller, A., Shuster, K., Urbanek, J.,
  Kiela, D., Szlam, A., Serban, I., Lowe, R., et~al.: The second conversational
  intelligence challenge (convai2).
\newblock arXiv e-prints pp. arXiv--1902 (2019)

\bibitem{fan2018hierarchical}
Fan, A., Lewis, M., Dauphin, Y.: Hierarchical neural story generation.
\newblock In: ACL (1) (2018)

\bibitem{grootendorst2020keybert}
Grootendorst, M.: Keybert: Minimal keyword extraction with bert. (2020).
\newblock \doi{10.5281/zenodo.4461265}.
\newblock \urlprefix\url{https://doi.org/10.5281/zenodo.4461265}

\bibitem{holtzman2019curious}
Holtzman, A., Buys, J., Du, L., Forbes, M., Choi, Y.: The curious case of
  neural text degeneration.
\newblock In: International Conference on Learning Representations (2019)

\bibitem{hsu2018emotionlines}
Hsu, C.C., Chen, S.Y., Kuo, C.C., Huang, T.H., Ku, L.W.: Emotionlines: An
  emotion corpus of multi-party conversations.
\newblock In: Proceedings of the Eleventh International Conference on Language
  Resources and Evaluation (LREC 2018) (2018)

\bibitem{li2016diversity}
Li, J., Galley, M., Brockett, C., Gao, J., Dolan, W.B.: A diversity-promoting
  objective function for neural conversation models.
\newblock In: Proceedings of the 2016 Conference of the North American Chapter
  of the Association for Computational Linguistics: Human Language
  Technologies, pp. 110--119 (2016)

\bibitem{li2016persona}
Li, J., Galley, M., Brockett, C., Spithourakis, G., Gao, J., Dolan, W.B.: A
  persona-based neural conversation model.
\newblock In: Proceedings of the 54th Annual Meeting of the Association for
  Computational Linguistics (Volume 1: Long Papers), pp. 994--1003 (2016)

\bibitem{li2018syntactically}
Li, J., Sun, X.: A syntactically constrained bidirectional-asynchronous
  approach for emotional conversation generation.
\newblock In: Proceedings of the 2018 Conference on Empirical Methods in
  Natural Language Processing, pp. 678--683 (2018)

\bibitem{li2020empdg}
Li, Q., Chen, H., Ren, Z., Ren, P., Tu, Z., Chen, Z.: Empdg: Multi-resolution
  interactive empathetic dialogue generation.
\newblock In: Proceedings of the 28th International Conference on Computational
  Linguistics, pp. 4454--4466 (2020)

\bibitem{li2020empathetic}
Li, Q., Li, P., Chen, Z., Ren, Z.: Towards empathetic dialogue generation over
  multi-type knowledge (2020)

\bibitem{li2017dailydialog}
Li, Y., Su, H., Shen, X., Li, W., Cao, Z., Niu, S.: Dailydialog: A manually
  labelled multi-turn dialogue dataset.
\newblock In: Proceedings of the Eighth International Joint Conference on
  Natural Language Processing (Volume 1: Long Papers), pp. 986--995 (2017)

\bibitem{lin2019moel}
Lin, Z., Madotto, A., Shin, J., Xu, P., Fung, P.: Moel: Mixture of empathetic
  listeners.
\newblock In: Proceedings of the 2019 Conference on Empirical Methods in
  Natural Language Processing and the 9th International Joint Conference on
  Natural Language Processing (EMNLP-IJCNLP), pp. 121--132 (2019)

\bibitem{lin2020caire}
Lin, Z., Xu, P., Winata, G.I., Siddique, F.B., Liu, Z., Shin, J., Fung, P.:
  Caire: An end-to-end empathetic chatbot.
\newblock In: Proceedings of the AAAI Conference on Artificial Intelligence,
  vol.~34, pp. 13,622--13,623 (2020)

\bibitem{DBLP:journals/corr/abs-1907-11692}
Liu, Y., Ott, M., Goyal, N., Du, J., Joshi, M., Chen, D., Levy, O., Lewis, M.,
  Zettlemoyer, L., Stoyanov, V.: Roberta: {A} robustly optimized {BERT}
  pretraining approach.
\newblock CoRR \textbf{abs/1907.11692} (2019).
\newblock \urlprefix\url{http://arxiv.org/abs/1907.11692}

\bibitem{loper2002nltk}
Loper, E., Bird, S.: Nltk: The natural language toolkit.
\newblock In: Proceedings of the ACL-02 Workshop on Effective Tools and
  Methodologies for Teaching Natural Language Processing and Computational
  Linguistics, pp. 63--70 (2002)

\bibitem{mohammad2018obtaining}
Mohammad, S.: Obtaining reliable human ratings of valence, arousal, and
  dominance for 20,000 english words.
\newblock In: Proceedings of the 56th Annual Meeting of the Association for
  Computational Linguistics (Volume 1: Long Papers), pp. 174--184 (2018)

\bibitem{naous2021empathetic}
Naous, T., Antoun, W., Mahmoud, R., Hajj, H.: Empathetic bert2bert
  conversational model: Learning arabic language generation with little data.
\newblock In: Proceedings of the Sixth Arabic Natural Language Processing
  Workshop, pp. 164--172 (2021)

\bibitem{peng2020few}
Peng, B., Zhu, C., Li, C., Li, X., Li, J., Zeng, M., Gao, J.: Few-shot natural
  language generation for task-oriented dialog.
\newblock In: Proceedings of the 2020 Conference on Empirical Methods in
  Natural Language Processing: Findings, pp. 172--182 (2020)

\bibitem{pennington2014glove}
Pennington, J., Socher, R., Manning, C.D.: Glove: Global vectors for word
  representation.
\newblock In: Proceedings of the 2014 conference on empirical methods in
  natural language processing (EMNLP), pp. 1532--1543 (2014)

\bibitem{radford2018improving}
Radford, A., Narasimhan, K., Salimans, T., Sutskever, I.: Improving language
  understanding by generative pre-training

\bibitem{radford2019language}
Radford, A., Wu, J., Child, R., Luan, D., Amodei, D., Sutskever, I.: Language
  models are unsupervised multitask learners.
\newblock OpenAI blog \textbf{1}(8), 9 (2019)

\bibitem{rashkin2018know}
Rashkin, H., Smith, E.M., Li, M., Boureau, Y.L.: I know the feeling: Learning
  to converse with empathy  (2018)

\bibitem{rashkin2019towards}
Rashkin, H., Smith, E.M., Li, M., Boureau, Y.L.: Towards empathetic open-domain
  conversation models: A new benchmark and dataset.
\newblock In: ACL (1) (2019)

\bibitem{rothe2020leveraging}
Rothe, S., Narayan, S., Severyn, A.: Leveraging pre-trained checkpoints for
  sequence generation tasks.
\newblock Transactions of the Association for Computational Linguistics
  \textbf{8}, 264--280 (2020)

\bibitem{sennrich2016neural}
Sennrich, R., Haddow, B., Birch, A.: Neural machine translation of rare words
  with subword units.
\newblock In: Proceedings of the 54th Annual Meeting of the Association for
  Computational Linguistics (Volume 1: Long Papers), pp. 1715--1725 (2016)

\bibitem{serban2016generative}
Serban, I.V., Lowe, R., Charlin, L., Pineau, J.: Generative deep neural
  networks for dialogue: A short review (2016)

\bibitem{serban2015hierarchical}
Serban, I.V., Sordoni, A., Bengio, Y., Courville, A., Pineau, J.: Hierarchical
  neural network generative models for movie dialogues.
\newblock arXiv preprint arXiv:1507.04808 \textbf{7}(8), 434--441 (2015)

\bibitem{shen2020cdl}
Shen, L., Feng, Y.: Cdl: Curriculum dual learning for emotion-controllable
  response generation.
\newblock In: Proceedings of the 58th Annual Meeting of the Association for
  Computational Linguistics, pp. 556--566 (2020)

\bibitem{skowron2013affect}
Skowron, M., Theunis, M., Rank, S., Kappas, A.: Affect and social processes in
  online communication--experiments with an affective dialog system.
\newblock IEEE Transactions on Affective Computing \textbf{4}(3), 267--279
  (2013)

\bibitem{speer2017conceptnet}
Speer, R., Chin, J., Havasi, C.: Conceptnet 5.5: An open multilingual graph of
  general knowledge.
\newblock In: Proceedings of the AAAI Conference on Artificial Intelligence,
  vol.~31 (2017)

\bibitem{sutskever2014sequence}
Sutskever, I., Vinyals, O., Le, Q.V.: Sequence to sequence learning with neural
  networks.
\newblock Advances in Neural Information Processing Systems \textbf{27},
  3104--3112 (2014)

\bibitem{vaswani2017attention}
Vaswani, A., Shazeer, N., Parmar, N., Uszkoreit, J., Jones, L., Gomez, A.N.,
  Kaiser, L., Polosukhin, I.: Attention is all you need.
\newblock In: NIPS (2017)

\bibitem{wei2019emotion}
Wei, W., Liu, J., Mao, X., Guo, G., Zhu, F., Zhou, P., Hu, Y.: Emotion-aware
  chat machine: Automatic emotional response generation for human-like
  emotional interaction.
\newblock In: Proceedings of the 28th ACM International Conference on
  Information and Knowledge Management, pp. 1401--1410 (2019)

\bibitem{wolf2019transfertransfo}
Wolf, T., Sanh, V., Chaumond, J., Delangue, C.: Transfertransfo: A transfer
  learning approach for neural network based conversational agents (2019)

\bibitem{young2018augmenting}
Young, T., Cambria, E., Chaturvedi, I., Zhou, H., Biswas, S., Huang, M.:
  Augmenting end-to-end dialogue systems with commonsense knowledge.
\newblock In: Proceedings of the AAAI Conference on Artificial Intelligence,
  vol.~32 (2018)

\bibitem{zandie2020emptransfo}
Zandie, R., Mahoor, M.H.: Emptransfo: A multi-head transformer architecture for
  creating empathetic dialog systems.
\newblock In: The Thirty-Third International Flairs Conference (2020)

\bibitem{zech2005talking}
Zech, E., Rim{\'e}, B.: Is talking about an emotional experience helpful?
  effects on emotional recovery and perceived benefits.
\newblock Clinical Psychology \& Psychotherapy: An International Journal of
  Theory \& Practice \textbf{12}(4), 270--287 (2005)

\bibitem{zhang2018personalizing}
Zhang, S., Dinan, E., Urbanek, J., Szlam, A., Kiela, D., Weston, J.:
  Personalizing dialogue agents: I have a dog, do you have pets too?
\newblock In: ACL (1) (2018)

\bibitem{zhang2020dialogpt}
Zhang, Y., Sun, S., Galley, M., Chen, Y.C., Brockett, C., Gao, X., Gao, J.,
  Liu, J., Dolan, W.B.: Dialogpt: Large-scale generative pre-training for
  conversational response generation.
\newblock In: Proceedings of the 58th Annual Meeting of the Association for
  Computational Linguistics: System Demonstrations, pp. 270--278 (2020)

\bibitem{zhong2019knowledge}
Zhong, P., Wang, D., Miao, C.: Knowledge-enriched transformer for emotion
  detection in textual conversations.
\newblock In: Proceedings of the 2019 Conference on Empirical Methods in
  Natural Language Processing and the 9th International Joint Conference on
  Natural Language Processing (EMNLP-IJCNLP), pp. 165--176 (2019)

\bibitem{zhou2018emotional}
Zhou, H., Huang, M., Zhang, T., Zhu, X., Liu, B.: Emotional chatting machine:
  Emotional conversation generation with internal and external memory.
\newblock In: Thirty-Second AAAI Conference on Artificial Intelligence (2018)

\bibitem{zhou2018mojitalk}
Zhou, X., Wang, W.Y.: Mojitalk: Generating emotional responses at scale.
\newblock In: Proceedings of the 56th Annual Meeting of the Association for
  Computational Linguistics (Volume 1: Long Papers), pp. 1128--1137 (2018)

\end{thebibliography}

\end{document}